\documentclass[letterpaper, 10 pt, conference]{ieeeconf}  % Comment this line out if you need a4paper

\IEEEoverridecommandlockouts                              % This command is only needed if 
\usepackage{graphicx} % for pdf, bitmapped graphics files
\usepackage{hyperref}
\usepackage[noadjust]{cite}

\usepackage{amsmath}
\usepackage{adjustbox}
\usepackage{multirow}
\usepackage{pifont}
\usepackage{hhline}
\usepackage{gensymb}

\usepackage{algorithm}
\usepackage[noend]{algpseudocode}
\algrenewcommand\algorithmicrequire{\textbf{Input:}}
\algrenewcommand\algorithmicensure{\textbf{Output:}}
\algrenewcommand\algorithmicindent{.5em}%

\makeatletter
\newcommand\fs@spaceruled{\def\@fs@cfont{\bfseries}\let\@fs@capt\floatc@ruled
  \def\@fs@pre{\vspace{0.5em}\hrule height.8pt depth0pt \kern2pt}%
  \def\@fs@post{\kern2pt\hrule\relax}%
  \def\@fs@mid{\kern2pt\hrule\kern2pt}%
  \let\@fs@iftopcapt\iftrue}
\makeatother

\usepackage{xcolor}

\makeatletter
\let\MYcaption\@makecaption
\makeatother
\usepackage[font=footnotesize]{subcaption}
\makeatletter
\let\@makecaption\MYcaption
\makeatother

\title{\LARGE \bf
DVM-SLAM: Decentralized Visual Monocular Simultaneous Localization and Mapping for Multi-Agent Systems
}

\author{Joshua Bird, Jan Blumenkamp, and Amanda Prorok% <-this % stops a space
\thanks{All authors are affiliated with Department of Computer Science and Technology, University of Cambridge. Email: \{jyjb2@cantab.ac.uk, jb2270@cam.ac.uk, asp45@cam.ac.uk\}}%
}

\begin{document}

\maketitle
\thispagestyle{empty}
\pagestyle{empty}

%%%%%%%%%%%%%%%%%%%%%%%%%%%%%%%%%%%%%%%%%%%%%%%%%%%%%%%%%%%%%%%%%%%%%%%%%%%%%%%%
\begin{abstract}

Cooperative Simultaneous Localization and Mapping (C-SLAM) enables multiple agents to work together in mapping unknown environments while simultaneously estimating their own positions. This approach enhances robustness, scalability, and accuracy by sharing information between agents, reducing drift, and enabling collective exploration of larger areas. In this paper, we present Decentralized Visual Monocular SLAM (DVM-SLAM), the first open-source decentralized monocular C-SLAM system. By only utilizing low-cost and light-weight monocular vision sensors, our system is well suited for small robots and micro aerial vehicles (MAVs). DVM-SLAM's real-world applicability is validated on physical robots with a custom collision avoidance framework, showcasing its potential in real-time multi-agent autonomous navigation scenarios. We also demonstrate comparable accuracy to state-of-the-art centralized monocular C-SLAM systems. We open-source our code and provide supplementary material online\footnote{\url{https://proroklab.github.io/DVM-SLAM}}.

\end{abstract}

%%%%%%%%%%%%%%%%%%%%%%%%%%%%%%%%%%%%%%%%%%%%%%%%%%%%%%%%%%%%%%%%%%%%%%%%%%%%%%%%
\section{Introduction}
Multi-robot systems are becoming increasingly common as automation continues to grow across a variety of sectors, including self-driving cars \cite{xu_cobevt_2023}, drone swarms \cite{vasarhelyi_optimized_2018}, \cite{corke_networked_2005}, logistics \cite{sharon_conflict-based_2015}, \cite{martinoli_modeling_2004}, and surveillance/search and rescue \cite{kolling_surveillance_2009}, \cite{ smith_distributed_2018}. Underlying foundational concepts such as flocking \cite{reynolds_flocks_1987}, formation control \cite{oh_survey_2015}, coverage \cite{zelinsky_planning_1993}, or trajectory deconfliction \cite{van_den_berg_reciprocal_2008} require the agents to understand the world around them as well as their peers' locations within that world. This task is typically achieved through technologies such as GNSS or motion capture setups, however, not all environments have access to these systems.

These are scenarios where multi-agent SLAM provides a compelling solution, as it allows for mapping of unfamiliar environments while maintaining awareness of the agents' positions. However, many existing multi-agent SLAM implementations are centralized \cite{7989445, schmuck2019ccm, schmuck2021covins}, requiring agents to maintain a reliable communication link with a central server to operate. This is an impractical constraint for many real-world systems, as it introduces a single-point-of-failure, greatly limiting its use cases. Furthermore, centralized solutions raise concerns about scalability.

To overcome these limitations, decentralized SLAM systems have emerged \cite{xu2022d, tian22tro_kimeramulti, lajoieSwarmSLAM, Lajoie2020DOORSLAM, cieslewski2017dataefficientdecentralizedvisualslam} that do not rely on a central management server, allowing the agents to be deployed in environments where network infrastructure may be lacking. Instead of utilizing a central node, the agents communicate peer-to-peer to enable relative localization and to build a shared map. These systems rely on complex and costly sensors such as LiDAR, RGB-D cameras, or stereo vision setups, which are more accurate, but unsuitable for small robots or cost-sensitive applications.

In this paper, we introduce Decentralized Visual Monocular SLAM (DVM-SLAM), a novel decentralized cooperative SLAM system optimized for monocular vision. Monocular vision systems are light-weight, inexpensive, and ideal for resource-constrained platforms such as micro aerial vehicles (MAVs). DVM-SLAM enables agents to robustly localize themselves, estimate relative poses, and cooperatively map unknown environments using only monocular vision in situations where localization and communication may be frequently lost.

The main contributions of this paper include:
\begin{itemize}
    \item The presentation of DVM-SLAM, the first open-source decentralized monocular C-SLAM system, with accuracy comparable to state-of-the-art centralized systems.
    \item A novel approach to the decentralized pose graph optimization problem, facilitated by incremental and asynchronous peer-to-peer map sharing.
    \item The evaluation of DVM-SLAM using publicly available datasets and real-world collision avoidance experiments.
\end{itemize}

\begin{figure}[t]
    \centering
    \medskip
    \includegraphics[trim=5cm 5cm 5cm 5cm, width=0.8\linewidth]{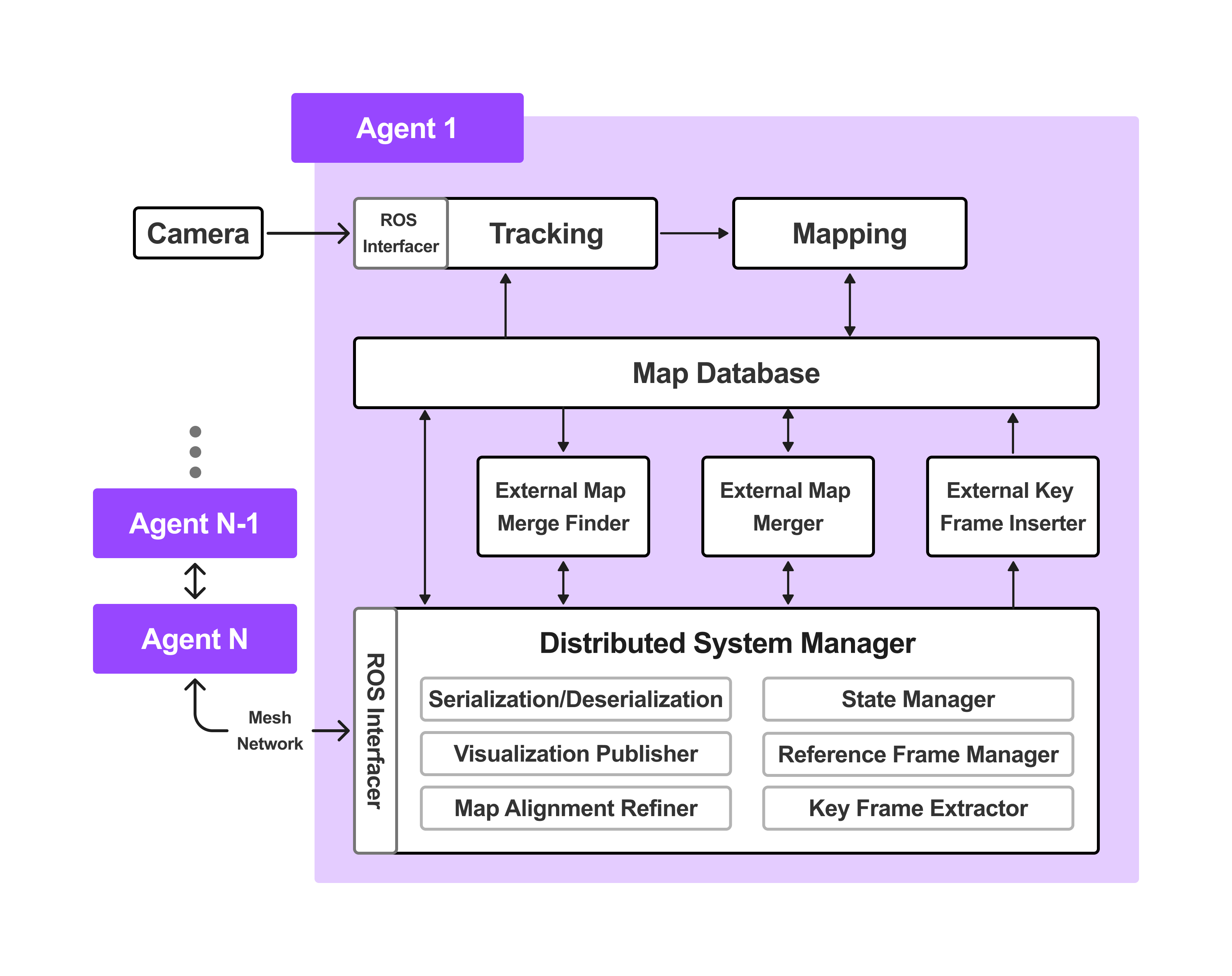}
    \caption{DVM-SLAM system architecture overview: each agent runs an instance of DVM-SLAM, which ingests camera data to localize the agent and build a map of the environment. The \textit{Distributed System Manager} interfaces with the \textit{Map Database} to merge maps with other agents, subsequently sharing keyframes and map points to cooperatively expand the shared map.}
    \label{fig:system-architecture}
\end{figure}

\section{Related Work}

\subsection{Decentralized Multi-Agent SLAM}
In recent years, there has been an emergence of decentralized multi-agent SLAM systems. DOOR-SLAM \cite{Lajoie2020DOORSLAM} leverages the Pairwise Consistency Maximization (PCM) method~\cite{8460217PCM} for outlier rejection in a distributed system. Kimera-multi \cite{tian22tro_kimeramulti} presented a distributed graduated non-convexity approach, demonstrating its superior performance to PCM. Swarm-SLAM \cite{lajoieSwarmSLAM} further advanced the field by introducing sparse inter-robot loop closure prioritization to reduce communication. However, as discussed in \autoref{sec:sensor-configuration}, none of these modern systems are capable of monocular vision operation.

Other systems such as \cite{doi:10.1126/scirobotics.abm5954, 8658783, DBLP:journals/corr/abs-2103-12770} require the agents to be initialized with their ground truth poses, which limits their real-world usability. In contrast, DVM-SLAM is able to provide accurate relative localization even when agents are initialized in arbitrary and unknown locations by identifying common landmarks in the world, as explained in \autoref{sec:map-merging}.

\subsection{Decentralized Pose Graph Optimization}
Pose graph optimization is the foundation of modern SLAM systems, as it allows for the refinement of a robot's estimated trajectory by optimizing the relative poses between frames, reducing drift, and improving the consistency of the map. In centralized multi-agent systems such as Schmuck et al.'s 2017 system \cite{7989445}, CCM-SLAM \cite{schmuck2019ccm}, and COVINS \cite{schmuck2021covins}, this task is performed by a central node and the resulting map is disseminated to all the agents. This requires the agents to maintain a permanent connection to the central node through which all data flows, which may be infeasible in real-world situations and scales poorly.

Decentralized systems such as DVM-SLAM present a more flexible and scalable solution, however the pose graph must be optimized without a central management node, which is known as the decentralized pose graph optimization (PGO) problem. There are various approaches to decentralized PGO. SWARM-SLAM \cite{lajoieSwarmSLAM} elects a single agent to perform the PGO for the entire swarm. This is simple, but similar to centralized systems, it scales poorly as only one robot is used to perform the computations for the entire swarm. Additionally, this method requires all agents to send their pose estimations before each optimization, resulting in additional communication overhead.

Other systems choose to perform PGO by spreading computation across agents through the use of distributed optimization frameworks. For example, $D^2$SLAM \cite{xu2022d} utilizes the ARock algorithm \cite{Peng_2016}, DOOR-SLAM \cite{Lajoie2020DOORSLAM} implements a two-stage distributed Gauss-Seidel method~\cite{DBLP:journals/corr/ChoudharyCNRCD17}, and Kimera-Multi \cite{tian22tro_kimeramulti} adopts a Riemannian block-coordinate descent method~\cite{tian2021distributedcertifiablycorrectposegraph}.

Instead of performing discrete optimization runs, our method of decentralized PGO is performed incrementally and asynchronously. Each agent optimizes its pose graph as external observations streams in, with a separate map alignment step. This method has no additional communication overhead, apart from the infrequent map alignment step, however, it comes at the cost of less verifiable global consistency. This method and its characteristics are further discussed in \autoref{sec:externally-generated-keyfrane-insertion} and subsequently evaluated in \autoref{sec:experimental-results}, demonstrating its strong real-world performance.

\subsection{Sensor Configuration}
\label{sec:sensor-configuration}
Many recent decentralized systems \cite{xu2022d, schmuck2021covins, tian22tro_kimeramulti, lajoieSwarmSLAM, Lajoie2020DOORSLAM} do not support monocular vision based sensors, as they introduce several challenges, including arbitrary map scale and more frequent loss of localization due to lower feature density. \autoref{fig:related-work-sensor-configurations} presents the sensor configurations used by popular state-of-the-art C-SLAM systems. 
%Recent single-agent SLAM systems \cite{greene2020metrically}, \cite{8100178} have found success in employing convolutional neural networks to accurately determine scale using purely monocular visual data.

\begin{table}[tb]
    \centering
    \medskip
    \caption{Comparison of popular C-SLAM system sensor configuration.}
    \label{fig:related-work-sensor-configurations}
    \adjustbox{valign=t, width=\linewidth}{
        \def\arraystretch{1.24}
\begin{tabular}{ |l|c|c|c|c|c|c|c|c|c| }
    \hline
    \multirow{2}{*}{\textbf{System}}                       & \textbf{Collaboration} & \multirow{2}{*}{\textbf{Monocular}} & \textbf{Monocular}     & \multirow{2}{*}{\textbf{Stereo}} & \textbf{Stereo}        & \textbf{LiDAR}         & \textbf{RGBD}          \\
                                                           &                      \textbf{Type}                     &                                                                       & \textbf{\texttt{+}IMU} & & \textbf{\texttt{+}IMU} & \textbf{\texttt{+}IMU} & \textbf{\texttt{+}IMU} \\

    \hhline{|=|=|=|=|=|=|=|=|=|=|}
    \textbf{DVM-SLAM [ours]}                                                                & Decentralized          & \ding{51} &                                                                    &                        &                        &                        &                        \\
    \hline
    $D^2$SLAM \cite{xu2022d}                                          & Decentralized          &                          &                                             &                        & \ding{51}                      &                        &                        \\
    \hline
    Kimera-Multi \cite{tian22tro_kimeramulti}                            & Decentralized          &      &                                                                 &                        & \ding{51}                      &                        & \ding{51}                      \\
    \hline
    Swarm-SLAM \cite{lajoieSwarmSLAM}                                    & Decentralized          &       &                                                                &                        & \ding{51}                      & \ding{51}                      & \ding{51}                      \\
    \hline
    DOOR-SLAM \cite{Lajoie2020DOORSLAM}                                  & Decentralized          &    &                                                                   &                        & \ding{51}                      &  \ding{51}                      &                        \\
    \hline
    Cieslewski et al. \cite{cieslewski2017dataefficientdecentralizedvisualslam}                                  & Decentralized          &                                                                       &                        &   \ding{51} &                    &                        &                        \\
    \hhline{|=|=|=|=|=|=|=|=|=|=|}
    CCM-SLAM \cite{schmuck2019ccm}                                      & Centralized            & \ding{51}        &                                                             &                        &                        &                        &                        \\
    \hline
    COVINS \cite{schmuck2021covins}                                      & Centralized            &                                                                       & \ding{51}     &                 &                        &                        &                        \\
    \hline
\end{tabular}

    }
\end{table}

While existing decentralized monocular visual SLAM systems exist \cite{egodagamage2017collaborative, bresson2012real, chen2018distributed}, they generally lack robust performance, with none presenting a quantitative evaluation of their system on publicly available datasets or providing open-source code. This is in contrast to DVM-SLAM, which is shown to perform on-par with centralized C-SLAM systems on public datasets and has been deployed in real-world experiments, in addition to being made open-source.

Therefore, DVM-SLAM stands out as a robust decentralized SLAM system optimized for monocular vision based sensors. This is advantageous for certain use cases, as LiDAR and RGB-D sensors have considerable weight, and stereo cameras may require a minimum camera separation to operate, both of which limit their usability on devices such as small aerial robots. Additionally, monocular camera sensors are ubiquitous and cheap, further enhancing their appeal for widespread deployment in resource-constrained platforms.

\section{System Overview}
DVM-SLAM's system architecture is given in \autoref{fig:system-architecture}. Each agent runs an instance of DVM-SLAM, ingesting camera data for agent localization and map-building using the \textit{Tracking} and \textit{Mapping} components respectively, which are taken from \cite{ORBSLAM3_TRO}. The novel \textit{Distributed System Manager} interfaces with the \textit{Map Database} to identify and perform map merges with other agents. After merging, the agents collaborate by sharing keyframes and map points to expand the shared map.

\begin{figure}[h]
    \centering
    \medskip
    \begin{subfigure}[b]{0.475\linewidth}
        \centering
        \includegraphics[trim=5cm 4cm 5cm 5cm, width=\linewidth]{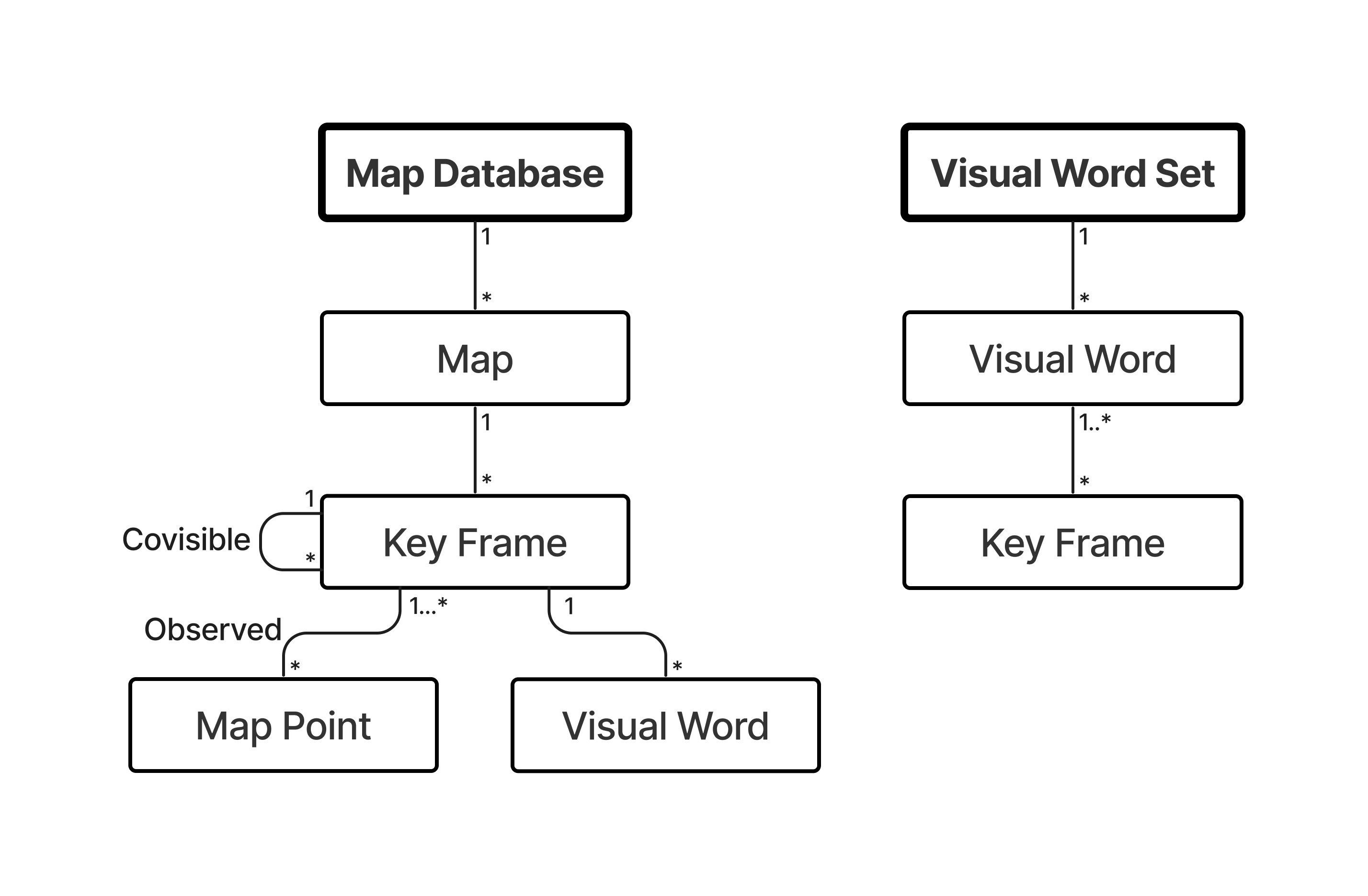}
    \end{subfigure}\hfill%
    ~
    \begin{subfigure}[b]{0.425\linewidth}
        \centering
        \includegraphics[width=\linewidth]{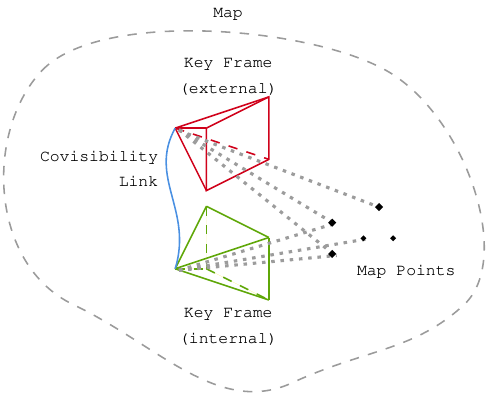}
    \end{subfigure}%
    \caption{The \textit{Map Database} structure represents the shared map through keyframes and map points, where each keyframe is additionally associated with the visual words it contains. Multiple maps may be held in the \textit{Map Database} in the case where the agent loses localization. The \textit{Visual Word Set} organizes keyframes by indexing them by their visual words. The right figure gives a visual representation of the \textit{Map Database}.}
    \label{fig:datastructure-diagram}
\end{figure}

\autoref{fig:datastructure-diagram} presents the \textit{Map Database} and \textit{Visual Word Set} data structures, with the latter being used to detect potential merges in \autoref{sec:map-merging}.

\subsection{Agent Communication Model}
We assume a system with $N$ agents $A=\{agent_1, agent_2, ..., agent_N\}$, where $agent_i$ is the agent with ID $i$. Within this system, we maintain a state $S_i=state_{i-j}$ for every agent pair $(agent_i, agent_j)$. We additionally maintain a set $G$ which contains all groups of agents, where a \textit{group} is defined as a collection of agents that are in communication with each other and have merged, therefore sharing the same coordinate frame. The \textit{group leader} is defined as the agent in a group with the lowest ID. It is important to note that the group leader is dynamically assigned; if the leader fails, a new one is automatically selected, preserving the system's decentralized property.

In the case where agents can lose communication with one another, we also assume that if any given $agent_i$ is able to communicate with an $agent_j$, $agent_i$ can also communicate with all of $agent_j$'s connected peers. This is held if the agents are using a mesh network to communicate.

Initially, all agents are in separate coordinate frames, so every state in $S$ is set as \textit{unmerged} and $G=\{\{agent_1\}, \{agent_2\}, ..., \{agent_N\}\}$.

Crucially, all merge operations are delegated to group leaders. This is significant, as the computational load and bandwidth requirements of merge operations scale proportionally to the square of the number of agents involved, and in swarming use cases the number of group leaders quickly drops to be much lower than the total number of agents. Additionally, having all merge operations performed by the group leader prevents potential race conditions introduced by communication latency within a group.

\begin{figure}[h]
    \centering
    \medskip
    \includegraphics[width=\linewidth]{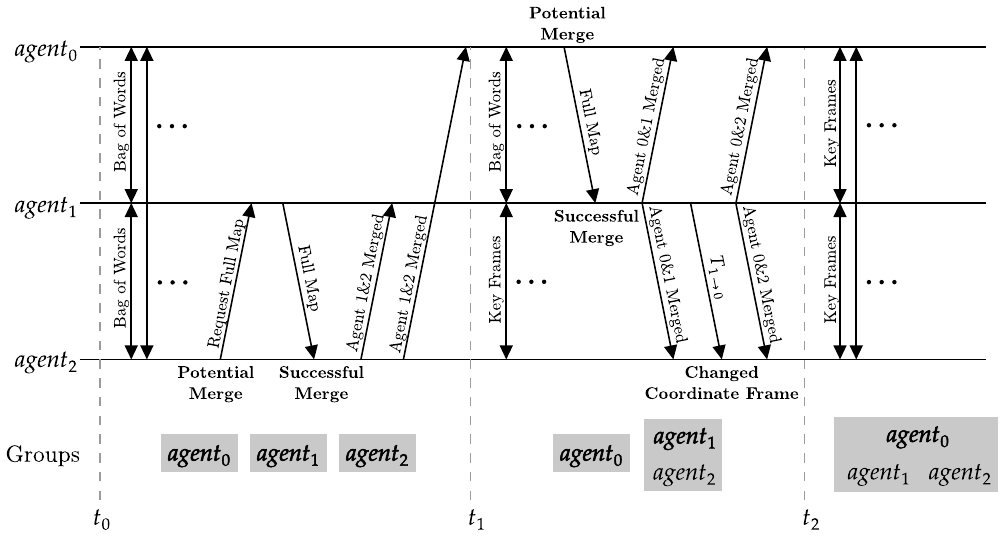}
    \caption{Messages exchanged in a simple 3 agent map merge example where $agent_1$ and $agent_2$ merge first, and then $agent_0$ and $agent_1$. Agents in the same group are shown in the same rectangle, with the group leader bolded. At $t_0$, all agents are unmerged. At $t_1$, $agent_1$ and $agent_2$ merge, and finally at $t_2$ all three agents are merged.}
    \label{fig:3-agent-merge}
\end{figure}

A full 3 agent merge example is given in \autoref{fig:3-agent-merge}.

\subsection{Decentralized System Manager} 
Decentralized SLAM systems are significantly more complex than single-agent or even centralized systems, due to the nuanced interactions between agents as they merge maps, lose localization, or lose connection with their peers. Therefore, a robust framework must be put in place to ensure the correctness of the system, which is handled by the \textit{Decentralized System Manager}.

\begin{figure}[b]
    \centering
    \includegraphics[trim=4cm 4cm 4cm 4cm, width=\linewidth]{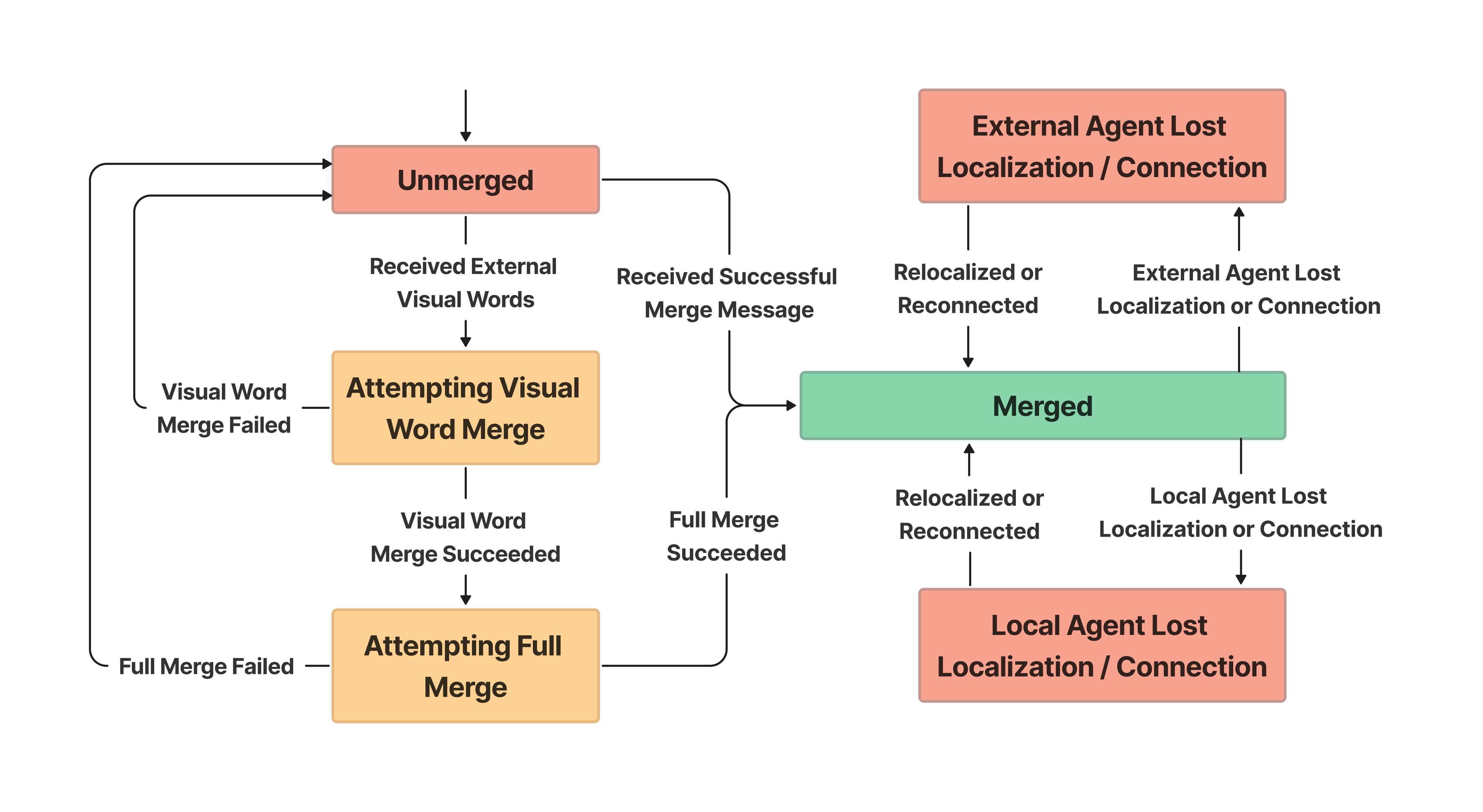}
    \caption{SLAM system state machine for a single agent, used to maintain the operations of the multi-agent system. Each transition between states is triggered by specific events, as illustrated in the diagram.}
    \label{fig:state-machine}
\end{figure}

Each agent's \textit{Decentralized System Manager} maintains a state machine for every peer in the system, shown in \autoref{fig:state-machine}. All peers are initialized in the \textit{unmerged} state, representing that they are in different coordinate frames. As an agent explores the same locations as its peers, the system recognizes the visual overlap and merges their maps, bringing them to the \textit{merged} state where the agents share the same coordinate frame and map, enabling relative positioning and cooperative map building.

\subsection{Map Merging}
\label{sec:map-merging}
As leader $agent_i$ generates new keyframes, it uses \cite{GalvezTRO12} to calculate the visual bag-of-words representation of the keyframes and sends them to all other leaders. Upon receiving the bag-of-words representation, the agent applies \autoref{alg:calculate-merge-score}, which leverages the spatial locality of its local map to assess the likelihood that the received bag-of-words corresponds to an area in its local map. It also returns the predicted \textit{BestMatchKF}, representing the location of the anticipated map merge. This score is then compared with a dynamic baseline merge score to determine if a merge is possible, generated by running \autoref{alg:calculate-merge-score} on the local map using the bag-of-words representation of \textit{BestMatchKF}. This process allows the bag-of-words-based merge detection to generalize across a diverse set of environments.

\floatstyle{spaceruled}% Select new float style
\restylefloat{algorithm}% Apply spaceruled float style to algorithm
\begin{algorithm}[t]
    \caption{Compute merge score between visual words and local map to find the best-matching keyframe.}
    \label{alg:calculate-merge-score}
    \begin{algorithmic}[1]
        \small
        \Procedure{CalculateMergeScore}{VisualWords}
        \State PotentialMatches $\gets$ query \textit{Visual Word Set} for keyframes sharing any visual words with input VisualWords
        \State BestMatchKF $\gets$ null
        \State BestMergeScore $\gets$ 0
        \For{each KF$_0$ in PotentialMatches}
        \State MergeScore $\gets$ KF$_0$ similarity to VisualWords
        \State Covisible $\gets$ 5 keyframess with highest covisibility with KF$_0$
        \For{each KF$_{cov}$ in Covisible} 
        \State MergeScore += KF$_{cov}$ similarity to VisualWords
        \EndFor
        \If{MergeScore $\geq$ BestMergeScore}
        \State BestMergeScore $\gets$ MergeScore
        \State BestMatchKF $\gets$ KF$_0$
        \EndIf
        \EndFor
        \State \Return (BestMergeScore, BestMatchKF)
        \EndProcedure
    \end{algorithmic}
\end{algorithm}

After a potential merge is found between $agent_i$ and some $agent_j$, where $i<j$ without loss of generality, $agent_i$ sends its full map to $agent_j$ which attempts a full map merge using all the data. If the merge is successful, $agent_j$ transforms its coordinate space to align with $agent_i$'s map using the transformation $T_{j \to i}$. Additionally, $agent_j$ sends $T_{j \to i}$ to the members of its group, allowing them to also transform into $agent_i$'s coordinate space. After this has been completed, set $S$ is updated according to \autoref{eq:state-update-equation} and $agent_i$ and $agent_j$'s groups in $G$ are merged.

{\small
\begin{equation}
    \begin{aligned} \label{eq:state-update-equation}
            \forall i \in g_n. \ \forall j \in g_m. \ state_{i-j} \in S \text{ and } state_{i-j} = \text{\textit{Merged}} \\ 
            \text{where $g_n, g_m \in G$ are groups led by $agent_n, agent_m$}
        \end{aligned}
\end{equation}
}

\subsection{Decentralized Pose Graph Optimization}
\label{sec:externally-generated-keyfrane-insertion}
DVM-SLAM performs incremental, asynchronous, and decentralized pose graph optimization through a keyframe sharing method. Agents within the same group operate in a common coordinate frame, allowing them to share their maps with one another. Each $agent_i$ maintains a set of unsent keyframes $K_\mathrm{unsent}$ and map points $M_\mathrm{unsent}$ for every other agent in its group. Once \#$K_\mathrm{unsent}$ exceeds a certain threshold, we serialize $K_\mathrm{unsent}$ and $M_\mathrm{unsent}$ and send them to the external agent. Finally, we set $K_\mathrm{unsent} = \emptyset$ and $M_\mathrm{unsent} = \emptyset$.

Upon receiving the serialized keyframes and map points, the agent deserializes them and appends them to the external keyframe queue to await insertion into their local copy of the shared map.

The \textit{External Keyframe Inserter} module is run whenever the agent has spare cycles on the CPU, to prevent it from impacting the local tracking and mapping performance. The insertion process involves the following operations:

\begin{enumerate}
    \item Pop $k_\mathrm{ext}$ from the front of the external keyframe queue.
    \item Move $k_\mathrm{ext}$ and its external observed map points $M_\mathrm{ext}$ to the local map. \\
        This can be performed without any transformations as $k_\mathrm{ext}$ and $M_\mathrm{ext}$ are in the same coordinate frame as the local map.
    \item Relink $k_\mathrm{ext}$ with co-visible keyframes and observed map points in the local map, and $M_\mathrm{ext}$ with local keyframes that observe them. \\
          $k_\mathrm{ext}$ and $M_\mathrm{ext}$ contain references to keyframes and map points that have already been sent or were generated by another agent. We search our local map for objects that match these references, reconnecting them in the \textit{Map Database}.
    \item Merge $M_\mathrm{ext}$ with map points in the local map. \\
          We exploit spatial locality to merge duplicate map points that describe the same physical feature. This ensures that local and external keyframes stay well connected, preventing map divergence.
    \item Perform a local pose graph optimization around $k_\mathrm{ext}$. \\
          This optimizes our map using the new information we received from the external agent.
\end{enumerate}

This decentralized PGO method offers several advantages, particularly in highly interactive multi-agent systems such as warehouse robots or drone swarms. By incrementally and asynchronously optimizing each agent’s pose graph as data is received, it is minimally impacted by network latency, agent disconnections, or infrequent communication, making it well suited for decentralized environments where network infrastructure may be limited. However, this method does require agents to communicate in a fully connected manner, which may be addressed through a decentralized hierarchical communication strategy.

% \begin{figure}[h]
%     \centering
%     \includegraphics[width=\linewidth]{figures/map_integration.png}

%     \caption{Section of the keyframes generated when running the EuRoC Machine Hall dataset. The local (green) and external (red) keyframes are well connected by co-visibility links (blue), demonstrating that they are tracking the same map points. \jan{This is a good caption too! Does this figure add value to the paper though?}}
%     \label{fig:well-connected-graph}
% \end{figure}

\subsection{Map Alignment Refiner}
As the shared map grows, an individual agent's map may ``fall out of alignment'', becoming slightly translated, rotated, or scaled with respect to the lead agent's map. This is largely a side effect of our early merge strategy which may merge two agents' maps before there is significant overlap, causing the estimated map alignment to have some error. These small alignment errors are acceptable when maps are small, but may cause the maps to diverge as they grow. 

To remedy this, we continuously refine our map alignment after merging. Map alignment is performed as follows:

\begin{enumerate}
    \item Request map point locations from the lead agent. This is defined as the set TaggedMP$_\mathrm{ext}$ where TaggedMP${_\mathrm{ext}}_i = (\text{uuid}, (x, y, z))$
    \item Extract local map point locations. This is defined as the set TaggedMP$_\mathrm{local}$ where TaggedMP${_\mathrm{local}}_i = (\text{uuid}, (x, y, z))$
    \item Use the Kabsch-Umeyama algorithm to find the SIM(3) transformation $T_\mathrm{local \to ext}$ from TaggedMP$_\mathrm{local}$ to TaggedMP$_\mathrm{ext}$, minimizing the root mean squared error. This is augmented by the RANSAC algorithm to reject outliers.
    \item Apply transformation $T_\mathrm{local \to ext}$ to our local map, realigning it with the group leader.
\end{enumerate}

We use an \textit{additive increase multiplicative decrease} methodology to control how often map alignment is performed. Given $t_i$ is the time between the $i$-th and $(i+1)$-th map alignments, we set $t_{i+1} = t_i + 1$ if the maps were well aligned, and $t_{i+1} = t_i / 2$ if the maps were not well aligned. This prevents agents from continuously performing map alignments if their maps are already well aligned.

\subsection{Losing Localization}
Recovering from a loss of localization is crucial to the robustness of a visual C-SLAM system in real-world scenarios where the environment may be lacking in texture and cameras may be temporarily obstructed. 
% \josh{This is particularly true in monocular systems where sensor data is less rich than to other types of sensors.}

If an agent loses localization within the shared map, it signals this to its peers and they stop exchanging keyframes. The agent then proceeds to build a new private map of the area it is observing. Using the same methods discussed in \autoref{sec:map-merging}, the agent is able to detect if it has revisited an area of the shared map and merges its private map with the shared map, finally signaling to its peers that it has regained localization. Keyframe sharing will then resume, with the agents' backlog of unsent keyframes being sent to one another, allowing the private map to become integrated into the shared map.

\section{Experimental Results}
\label{sec:experimental-results}

\begin{figure}[tb]
    \centering
    \medskip
    \includegraphics[width=0.95\linewidth]{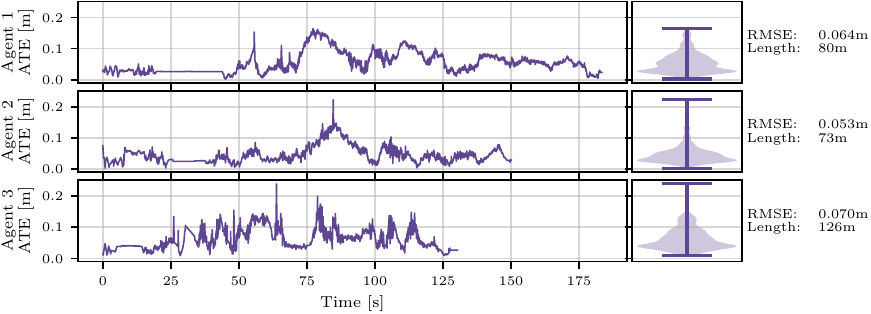}

    \caption{Plot of absolute trajectory error with respect to the ground truth when running the EuRoC Machine Hall 01-03 scenarios in parallel on three agents. The RMS absolute trajectory error is 6.2cm over the combined trajectory of 279 meters.}
    \label{fig:euroc-mh-01-03-line-plot}
\end{figure}

\begin{figure}[b]
    \centering
    \begin{subfigure}[b]{0.55\linewidth}
        \centering
        \marginbox{0 0 0 0} {
            \includegraphics[width=0.95\linewidth]{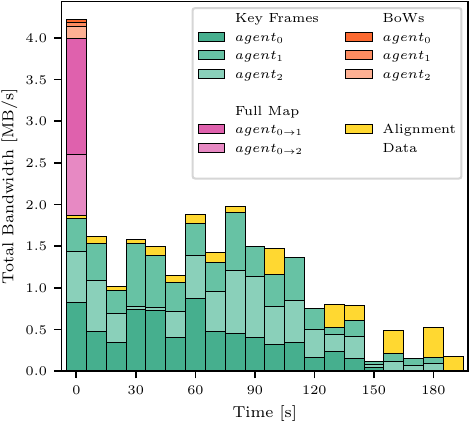}
        }
        \caption{Total system data over time, segregated by message type.}
    \end{subfigure}%
    ~
    \begin{subfigure}[b]{0.45\linewidth}
        \centering
        \adjustbox{valign=t, width=\linewidth}{
            \marginbox{0.2in 0 0 0} {
                
                \def\arraystretch{1.2}
                \begin{tabular}{ |c|l|r|r| }
                    \cline{3-4}
                    \multicolumn{2}{}{}                       & \multicolumn{1}{|c|}{KB} & \multicolumn{1}{|c|}{Avg. KB/s}         \\
                    \hline
                    \multirow{3}{*}{Key Frames}               & $agent_0$                & 69,971                          & 351.9 \\
                                                              & $agent_1$                & 63,908                          & 321.4 \\
                                                              & $agent_2$                & 65,164                          & 327.7 \\
                    \hline
                    \multirow{3}{*}{BoWs}                     & $agent_0$                & 371                             & 1.9   \\
                                                              & $agent_1$                & 437                             & 2.2   \\
                                                              & $agent_2$                & 1,496                           & 7.5   \\
                    \hline
                    \multirow{2}{*}{Full Map}
                                                              & $agent_{0\to1}$          & 13,953                          & 70.2  \\
                                                              & $agent_{0\to2}$          & 7,319                           & 36.8  \\
                    \hline
                    \multicolumn{2}{|c|}{Alignment Data}      & 22,560                   & 113.6                                   \\
                    \hline
                    \multicolumn{2}{|c|}{\textbf{Total Data}} & \textbf{245,218}         & \textbf{1,233.1}                        \\
                    \hline
                \end{tabular}
            }
        }
        \caption{Total system data by message type.}
        \vfill

        \adjustbox{valign=b, width=\linewidth}{
            \marginbox{0.2in 0 0 0.3in} {
                \def\arraystretch{1.2}
                \begin{tabular}{ |l|r|r|r|r| }
                    \cline{2-5}
                    \multicolumn{1}{}{} & \multicolumn{2}{|c|}{Sent} & \multicolumn{2}{|c|}{Received}                                                               \\
                    \cline{2-5}
                    \multicolumn{1}{}{} & \multicolumn{1}{|c|}{KB}   & \multicolumn{1}{|c|}{Avg. KB/s} & \multicolumn{1}{|c|}{KB} & \multicolumn{1}{|c|}{Avg. KB/s} \\
                    \hline
                    $agent_0$           & {114,585}           & {576.2}                  & {131,005}         & {658.8}                  \\
                    \hline
                    $agent_1$           & {64,345}            & {323.6}                  & {173,554}         & {872.8}                  \\
                    \hline
                    $agent_2$           & {66,660}            & {335.2}                  & {164,606}         & {827.7}                  \\
                    \hline
                \end{tabular}
            }
        }
        \caption{Data by agent.}
    \end{subfigure}%

    \caption{Bandwidth transmitted between agents during the EuRoC Machine Hall 01-03 scenarios. DVM-SLAM operates by exchanging various types of messages, including \textit{bag-of-words} (BoWs) to detect map merges, \textit{full maps} to perform a merge, \textit{keyframes} to incrementally share map data, and \textit{alignment data}.}
    \label{fig:euroc-mh-01-02-bandwith}
\end{figure}

\subsection{Dataset Experiments}
This section will benchmark DVM-SLAM's performance on industry-standard visual SLAM datasets. All dataset evaluations are run 5 times using only monocular camera data. We evaluate accuracy by calculating the Absolute Trajectory Error (ATE) metric \cite{6385773} of the combined agent trajectory.

% \begin{figure}[h]
%     \centering
%     \begin{subfigure}[t]{0.5\linewidth}
%         \includegraphics[width=\linewidth]{figures/apr11_mh_trajectory_b_trajectory.pdf}
%         \caption{EuRoC Machine Hall 01-03}
%         \label{fig:euroc-traj}
%     \end{subfigure}\hfill%
%     ~
%     \begin{subfigure}[t]{0.46\linewidth}
%         \includegraphics[width=\linewidth]{figures/apr11_tum_room_trajectory_a_trajectory.pdf}
%         \caption{TUM-VI Rooms 01-03}
%         \label{fig:tum-traj}
%     \end{subfigure}

%     \caption{Estimated trajectory and ground truth.}
% \end{figure}

\textbf{EuRoC Machine Hall}. This dataset \cite{burri2016euroc} provides a 752x480 20fps video feed and ground truth data at 20hz. The camera rig is attached to an aerial vehicle which flies through a 15x15 meter machine room. We synthesize a multi-agent dataset by concurrently running the Machine Hall 01-03 scenarios on three agents offline.

DVM-SLAM achieves an RMS absolute trajectory error of 5.9cm over the 279-meter total trajectory length, with a 1.4cm spread around the median across 5 trials. The median data transfer rate is 1.3MB/s with a spread of 0.12MB/s.

To further analyse the characteristics of DVM-SLAM, we focus on an individual trial. \autoref{fig:euroc-mh-01-03-line-plot} plots the ATE as the trial progresses, showing that the ATE returns to the baseline at the end of the trial when the agents return to their starting positions, with no perceived accumulated drift. This demonstrates the system's high global accuracy and relative positioning throughout the trial.

We now analyse the network usage presented in \autoref{fig:euroc-mh-01-02-bandwith}. Initially, the agents send bag-of-word information before quickly detecting a merge opportunity. The agents exchange their maps, which can be seen in the large initial spike in network bandwidth. After a successful merge, they begin exchanging keyframes. The rate of keyframe data fluctuates depending on how much new area the agents are exploring.

We also observe the agents sporadically sending alignment data to improve map consistency. This occurs less frequently the longer system runs due to the additive increase multiplicative decrease method used to schedule map alignments. However, the size of the messages also grows over time due to the growing map.

\begin{figure}[tb]
    \centering
    \medskip
    \includegraphics[width=0.95\linewidth]{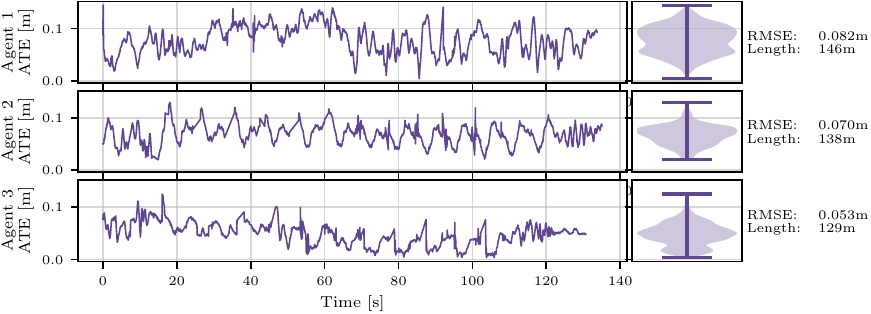}

    \caption{Plot of absolute trajectory error with respect to the ground truth when running the TUM-VI Room 1-3 scenarios in parallel on three agents. The RMS absolute trajectory error is 7.0cm over the combined trajectory of 413 meters.}
    \label{fig:tum-room-01-03-line-plot}
\end{figure}

% placed here for formatting
\begin{figure*}[ht]
    \centering
    \medskip
    \includegraphics[width=0.85\linewidth]{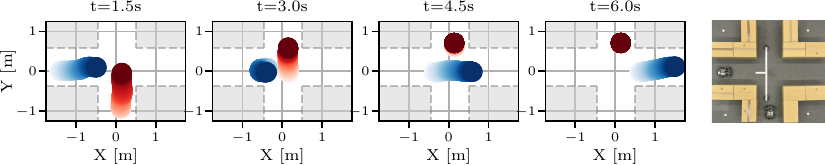}

    \caption{Demonstration of multi-agent collision avoidance. Two robots are set 90\degree{} to each other in an intersection environment, with no direct view of the other robot and little visual overlap (right). The blue agent is given a goal pose on the other side of the intersection and successfully avoids a collision when the red agent is pushed through the intersection. The trajectories generated by the SLAM system are presented on the left charts.}
    \label{fig:collision-avoidance}
\end{figure*}

\textbf{TUM-VI Rooms}. This dataset \cite{8593419} consists of handheld fisheye 512x512 video with ground truth data. The ``room'' environment is used for this evaluation, which is a 3x3 meter motion capture lab. There is less texture in this environment than the machine hall, making visual-only SLAM difficult. During the session, parts of the room are revisited dozens of times by different agents from different perspectives, allowing us to evaluate our system's ability to re-localize agents within previously mapped environments. Sessions 1-3 are combined to create an offline multi-agent dataset.

Across the 5 trials, DVM-SLAM achieved a RMS ATE of 6.95cm and average data transfer rate of 0.84MB/s with a spread of 0.58cm and 0.19MB/s respectively. 

Once again, we focus on an individual trial to further evaluate the system's performance. \autoref{fig:tum-room-01-03-line-plot} shows that there is no long-term error built up, demonstrating that DVM-SLAM is successfully localizing agents within the shared map and performing long-term map point association.

The data transfer characteristics are very similar to the results seen in the EuRoC Machine Hall experiment, and therefore yield similar conclusions. The only difference of note is a lower keyframe and alignment data transfer rate (236.3 KB/s and 49.0 KB/s, respectively) primarily due to the smaller environment.

\subsection{Comparison to Related Works}
\autoref{fig:comparison-to-multi-agent-systems} presents a comparison of DVM-SLAM to similar monocular C-SLAM systems, showcasing its performance in terms of RMS ATE across the EuRoC Machine Hall and TUM-VI Rooms datasets. CCM-SLAM is a centralized system, significantly simplifying their multi-agent SLAM problem compared to our decentralized system. For further comparison, the single-agent visual-inertial VINS-Mono system is included by using its multisession abilities, where the sessions are run consecutively with each session building upon the map built by the previous.

\begin{table}[b]
    \centering
    \caption{RMS absolute trajectory error of DVM-SLAM relative to centralized C-SLAM systems (median of 5 trials).}
    \def\arraystretch{1.2}
    \adjustbox{valign=b, width=\linewidth}{
    \begin{tabular}{ |c|c|c|c| }
        \cline{2-4}
        \multicolumn{1}{c|}{}    & DVM-SLAM & CCM-SLAM & VINS-Mono \\
        \hline
        EuRoC Machine Hall 01-03 & \textbf{0.059}     & 0.077        & 0.074         \\
        \hline
        Tum-VI Rooms 1-3         & \textbf{0.070}     & —       & 0.256         \\
        \hline
    \end{tabular}
    }

    \label{fig:comparison-to-multi-agent-systems}
\end{table}

In all cases, DVM-SLAM outperforms the other systems. Notably, CCM-SLAM failed all 5 trials of the TUM-VI Rooms dataset, due to its inability to re-localize an agent when localization is lost. DVM-SLAM does not suffer from this issue, demonstrating its robust tracking and ability to recover from losses of localization.

\subsection{Real-World Experiments}
Real-world experiments were performed to validate DVM-SLAM's real-time performance in a realistic environment with practical computational and communication constraints. Testing was conducted on the Cambridge RoboMaster platform \cite{blumenkamp2024cambridge}, equipped a NVidia Jetson Orin NX computer and Raspberry Pi HQ camera with a fisheye lens. Images were captured at a resolution of 960x540 at 15fps. DVM-SLAM was deployed on the robots along with a custom motion controller module which employs a non-linear model predictive control system to prevent collisions with static and dynamic obstacles.

\autoref{fig:collision-avoidance} tests the system in an intersection environment, where the two robots would normally collide. The agents are able to localize each other even when their views do not overlap and they can not see each other, demonstrating that a shared map is being built. Out of the four consecutive trials run in this environment, there were zero collisions between the two agents, highlighting the system's real-time performance in latency sensitive applications. The RMS ATE of the system was 7.4cm over the 50-meter-long trajectory.

% \begin{figure}[h]
%     \centering
%     \includegraphics[width=\linewidth]{figures/mar25_1_distance_plot.pdf}

%     \caption{Plot of distance between the two robots throughout all four collision avoidance trials. The dips between trials are the robots' positions being reset.}
%     \label{fig:collision-avoidance-distance-plot}
% \end{figure}

\section{Conclusion and Future Works}

In this paper, we presented DVM-SLAM, the first open-source decentralized visual monocular C-SLAM system. Our system allows multiple agents to coordinate in real-time, building shared maps of unknown environments while maintaining robust relative localization without relying on a centralized server. By leveraging monocular vision, DVM-SLAM addresses the size, weight, and cost constraints which often exist in practical deployments of multi-agent systems, such as a swam of MAVs, where LiDAR or stereo vision sensors may be impractical to use.

A novel aspect of our approach is the incremental decentralized pose graph optimization method, which allows agents to update their pose graphs as data becomes available, rather than relying on periodic, discrete optimization steps. This method minimizes communication overhead, making it well-suited for environments with limited network reliability. This approach does present challenges however, as there is a risk of map divergence in scenarios where there is minimal overlap between agent maps. Therefore, it is best suited for high interaction multi-agent scenarios where multiple agents are frequently visiting the same areas, such as long-lived indoor warehouse localization or MAV swarms operating in close formation.

Future work should focus on integrating inertial data to improve robustness in low-feature environment. Additionally, inertial data or learning-based approaches may be implemented to disambiguate map scale, a key limitation of monocular based SLAM systems. 

% Additionally, the bandwidth requirements of DVM-SLAM scale

\section*{Acknowledgements}
This work was supported in part by European Research Council (ERC) Project 949949 (gAIa). We also acknowledge a gift from Arm.

% \addtolength{\textheight}{-12cm}   % This command serves to balance the column lengths
                                  % on the last page of the document manually. It shortens
                                  % the textheight of the last page by a suitable amount.
                                  % This command does not take effect until the next page
                                  % so it should come on the page before the last. Make
                                  % sure that you do not shorten the textheight too much.

%%%%%%%%%%%%%%%%%%%%%%%%%%%%%%%%%%%%%%%%%%%%%%%%%%%%%%%%%%%%%%%%%%%%%%%%%%%%%%%%

%%%%%%%%%%%%%%%%%%%%%%%%%%%%%%%%%%%%%%%%%%%%%%%%%%%%%%%%%%%%%%%%%%%%%%%%%%%%%%%%

%%%%%%%%%%%%%%%%%%%%%%%%%%%%%%%%%%%%%%%%%%%%%%%%%%%%%%%%%%%%%%%%%%%%%%%%%%%%%%%%

%%%%%%%%%%%%%%%%%%%%%%%%%%%%%%%%%%%%%%%%%%%%%%%%%%%%%%%%%%%%%%%%%%%%%%%%%%%%%%%%

\bibliographystyle{IEEEtran}
\bibliography{ref}

\end{document}